\documentclass[journal,twoside,web]{ieeecolor}
\usepackage{color}
\usepackage{jsen}
\usepackage{jsen_compat}
\usepackage{cite}
\usepackage{amsmath,amssymb,amsfonts}
\usepackage{graphicx}
\usepackage{booktabs,array,tabularx}
\usepackage{wrapfig}
\usepackage[hidelinks]{hyperref}
\graphicspath{{figures/}}
\newcolumntype{Y}{>{\raggedright\arraybackslash}X}
\newcommand{\R}{\mathbb{R}}
\newcommand{\SE}{\mathrm{SE}}
\newcommand{\Exp}{\operatorname{Exp}}
\newcommand{\Log}{\operatorname{Log}}
\definecolor{abstractbg}{rgb}{0.89804,0.94510,0.83137}
\hypersetup{pdftitle={Spacecraft Pose Tracking via Kinematic Prediction and Selective Event Correction},pdfauthor={Yonglong Zhang; Guanning Yao; Shuo Zhang; Yifan Yang; Yang Liu; Zongwu Xie}}

\begin{document}
\title{Spacecraft Pose Tracking via Kinematic Prediction and Selective Event Correction}
\author{\mbox{Yonglong Zhang}, \mbox{Guanning Yao}, \mbox{Shuo Zhang},
\mbox{Yifan Yang}, \mbox{Yang Liu}\textsuperscript{\ensuremath{\dagger}}, and \mbox{Zongwu Xie}%
\thanks{\textsuperscript{\ensuremath{\dagger}}Corresponding author: Yang Liu (e-mail: \href{mailto:liuyanghit@hit.edu.cn}{liuyanghit@hit.edu.cn}).}%
\thanks{This work was supported by the Natural Science Foundation of Heilongjiang Province for Excellent Young Scholars under Grant YQ2024E018.}%
\thanks{All authors are with the State Key Laboratory of Robotics and Systems, Harbin Institute of Technology, Harbin 150001, Heilongjiang, China.}\endgraf}

\IEEEtitleabstractindextext{%
\fcolorbox{abstractbg}{abstractbg}{%
\begin{minipage}{\textwidth}%
\begin{wrapfigure}[18]{r}{2.85in}
\vspace{-6pt}
\includegraphics[width=2.85in]{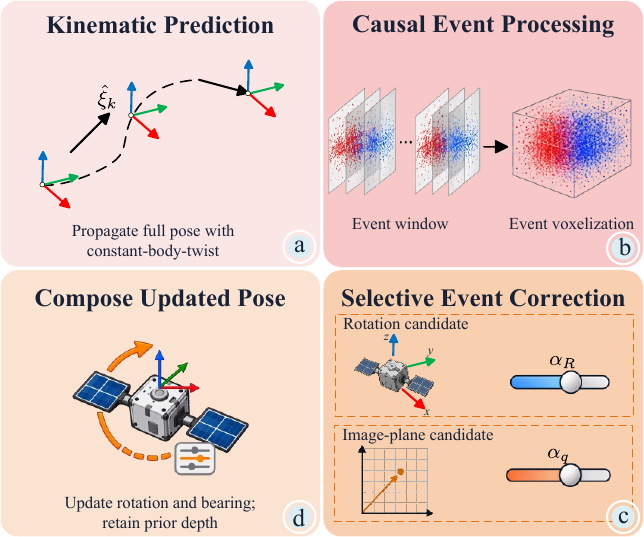}
\end{wrapfigure}
\begin{abstract}
In initialized spacecraft tracking, a short-horizon motion prediction remains available as event support changes. Observation-Aligned Event Innovation (OA-EI) uses events to correct selected components of this prediction. A constant-body-twist model propagates the complete pose; the event encoder proposes rotational and normalized image-plane translation corrections, retaining propagated axial depth. Candidate magnitudes come from a bounded Scale Gate and a deterministic activity factor, with empty windows returning the prediction unchanged. Evaluation uses four FRESH physical-event trajectories, twenty rollout starts, and three training seeds. Each rollout starts from two reference poses and then uses self-fed estimates. At 7.7 seconds, rotation and translation errors fall from $41.63^\circ$/57.28 mm for prediction alone to $15.73^\circ$/22.41 mm for OA-EI, reductions of 62.2\% and 60.9\%. Failure rates fall from 20.0\% to 1.67\%. Matched ablations favor the rotation--bearing interface for overall pose accuracy; combining scaling with activity support gives the lowest errors at every tested event-retention level. On Jetson Orin Nano, the measured in-memory update takes 44.56 milliseconds at the median and 58.77 milliseconds at P95.
\end{abstract}
\end{minipage}}}
\maketitle

\section{Introduction}
\IEEEPARstart{E}{vent} cameras record brightness changes asynchronously, offering fine temporal sampling and a wide dynamic range \cite{gallego2022survey}. For spacecraft tracking, the projected structure changes with motion and illumination, and event support varies in both amount and spatial coverage. Relative navigation, inspection, and robotic servicing require continuous estimates of target orientation and translation through these changes. Sparse observations can coincide with motion that is still predictable over a short interval.

Most spacecraft perception pipelines recover pose from current image evidence. SPEED and SPEED+ benchmark image-based estimation \cite{kisantal2020speed,park2022speedplus}; SPADES and FRESH extend evaluation to synthetic and physical event recordings \cite{rathinam2024spades,jawaid2026fresh}. Event-based approaches recover geometric keypoints for pose estimation, then use smoothing or optimization to improve temporal consistency \cite{rathinam2026onboard,zeng2026smoothing,zhang2026cascaded}. Initialized tracking also has recent pose history. A short-horizon motion model can predict the complete next state before the current event window is used for correction.

With that prediction available, the event branch must decide which components to correct and how much correction to inject. Full-pose re-estimation includes components already represented by the motion model; a selective update leaves the remaining components to the prediction. The magnitude matters beyond the current timestamp. An inaccurate correction enters the pose history used for later predictions, coupling state selection with update scaling throughout the rollout.

Observation-Aligned Event Innovation (OA-EI) applies this selective update to a complete six-degree-of-freedom prior from constant-body-twist propagation. The event branch predicts rotational and normalized image-plane translation candidates relative to the prior, while axial depth stays with propagation. Translation is thus split into bearing correction and depth prediction. Two scale coefficients, learned from error-minimizing scales along frozen candidates, set the correction magnitudes. Event count and spatial activity supply a deterministic support factor. The corrected pose becomes part of the next prediction history; with an empty window, the current prior is returned exactly. Neither CAD geometry nor keypoints are required by the event branch.

Prediction alone, full candidate injection, and OA-EI are evaluated under a common recursive protocol. Matched state-update and injection ablations examine state selection and correction scaling separately from this main comparison.

The contributions are:
\begin{itemize}
\item The event-to-pose interface corrects rotation and bearing while retaining axial depth from the complete kinematic prediction.
\item Candidate-scale supervision is combined with event-activity support to bound injection. Empty event windows give an exact prediction fallback.
\item Multi-start self-fed evaluation on physical event recordings shows reduced recursive pose errors and improved accuracy under severe event sparsification. Matched ablations examine state selection and injection scaling.
\end{itemize}

\section{Related Work}
\subsection{Spacecraft Pose Sensing}
Direct pose regression and image keypoints followed by perspective-$n$-point (PnP) geometry are the main routes to monocular spacecraft pose estimation \cite{pauly2023survey,sharma2020spn,lepetit2009epnp}. Both recover the complete pose from an image, but keypoint pipelines retain an explicit link to known object geometry rather than learning the whole image-to-pose mapping. SPEED and SPEED+ benchmark these approaches in synthetic and laboratory domains \cite{kisantal2020speed,park2022speedplus}.

For event-based estimation, SPADES provides synthetic sequences \cite{rathinam2024spades}, and FRESH supplies aligned physical event and RGB recordings under harsh illumination \cite{jawaid2026fresh}. FRESH's event-only baseline forms keypoint measurements for PnP; its full fusion method combines event and RGB evidence. Compact event-frame keypoint networks have also been coupled with neuromorphic hardware by Rathinam et al. \cite{rathinam2026onboard}. Temporal refinement takes different forms: PASAN and IRIS use polarity-aware temporal features with kinematic smoothing \cite{zeng2026smoothing}, and the cascaded method in \cite{zhang2026cascaded} adds geometric pose optimization and a motion-constrained backend. OA-EI starts its event update from the historical pose prediction, treating that prediction as the state to refine.

\subsection{Event-Based Motion Processing}
Grid representations allow convolutional networks to process asynchronous events \cite{gehrig2019representation}. Other motion-processing approaches work through event alignment, as in contrast maximization \cite{gallego2018contrast}, or image displacement, as in motion-compensated optical flow \cite{zhu2019flow}. Geometric odometry \cite{rebecq2017evo,zhou2021esvo} and learned odometry \cite{klenk2024devo} recover camera motion from event streams. Event6D instead tracks novel objects in six degrees of freedom using events and depth \cite{kang2026event6d}. For target-relative correction, OA-EI uses polarity-separated voxels and normalized image-plane translation, with axial depth supplied by the pose predictor.

\subsection{Prediction and Observation Updates}
Adaptive unscented Kalman filters for spacecraft tracking fuse neural pose measurements with motion predictions and adjust noise models or measurement uncertainty \cite{park2023ukf,cassinis2023uncertainty}. KalmanNet introduces recurrent learning into the filter when dynamics are partially known \cite{revach2022kalmannet}; latent Kalman filtering performs spacecraft pose prediction and learned observation fusion in feature space \cite{park2026drift}. These estimators combine prediction and observation to address measurement noise and model mismatch. OA-EI places its learned correction in a selected pose subspace: rotational and bearing-space candidates have separate scale coefficients, and event activity supplies an additional support factor.

\begin{figure}[!t]
\centering
\includegraphics[width=\columnwidth]{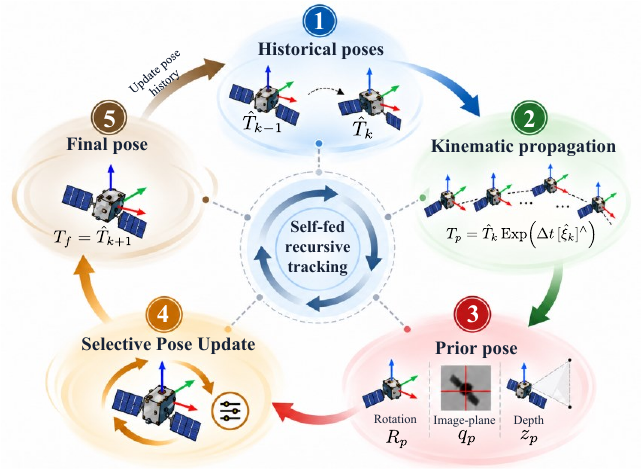}
\caption{Self-fed pose-tracking loop. Historical estimates are propagated into the prior $T_p=(R_p,t_p)$, represented by rotation $R_p$, image-plane coordinates $q_p$, and axial depth $z_p$. Selective correction yields $T_f=\hat T_{k+1}$; the next history is $(\hat T_k,\hat T_{k+1})$.}
\label{fig:framework}
\end{figure}

\section{Method}
Given $\hat T_{k-1},\hat T_k$ and a causal event window ending at $t_{k+1}$, OA-EI forms a complete prediction $T_p$ and applies rotational and image-plane corrections to obtain $T_f=\hat T_{k+1}$ (Fig.~\ref{fig:framework}). The corrected pose enters the next history. An \emph{event innovation} is the learned correction relative to $T_p$; its injected magnitude depends on learned scaling and current event activity.

\subsection{Kinematics-Guided Pose Prior}
Let $T_k={}^CT_{O,k}\in\SE(3)$ map the fixed object frame $O$ to the camera frame $C$:
\begin{equation}
T_k=\begin{bmatrix}R_k&t_k\\0^\top&1\end{bmatrix},
\qquad t_k=[x_k,y_k,z_k]^\top .
\end{equation}
In FRESH's fixed-camera setting \cite{jawaid2026fresh}, $C$ is constant within each sequence. With $\Delta t_j=t_{j+1}-t_j$, the body-frame twist and right-multiplicative prediction are
\begin{align}
\hat\xi_k&=\frac{1}{\Delta t_{k-1}}
 \Log(\hat T_{k-1}^{-1}\hat T_k)^\vee,\\
T_p&=\hat T_k\Exp(\Delta t_k[\hat\xi_k]^\wedge).
\label{eq:prior}
\end{align}
The wedge and vee operators map between twist vectors and Lie-algebra matrices. Write $T_p=(R_p,t_p)$ and $t_p=[x_p,y_p,z_p]^\top$. Propagation uses the historical motion trend; unmodeled acceleration and accumulated history errors motivate correction from current events.

\begin{figure}[!t]
\centering
\includegraphics[width=\columnwidth]{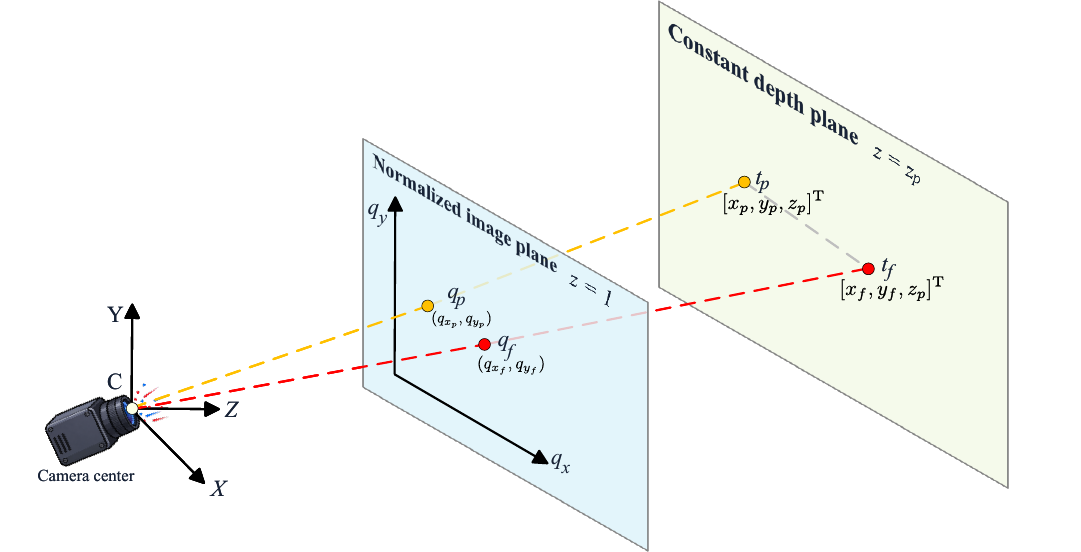}
\caption{Geometry of selective translation correction. The prior $t_p$ projects to $q_p=[x_p/z_p,y_p/z_p]^\top$. The event update changes $q_f=q_p+\alpha_q\hat{\delta q}$ and retains $z_f=z_p$, giving $t_f=z_p[q_{x,f},q_{y,f},1]^\top$. Both translations lie on the same axial-depth plane.}
\label{fig:geometry}
\end{figure}

\subsection{Observation-Aligned Correction}
The causal event window contains pixel locations $(u_i,v_i)$, timestamps $\tau_i$, and polarities $p_i$:
\begin{equation}
\mathcal E_{k+1}^{(h)}=
 \{(u_i,v_i,\tau_i,p_i)\mid t_{k+1}-h<\tau_i\le t_{k+1}\}.
\end{equation}
Here $h$ is the observation history and $\Delta t_k$ the propagation interval. Polarity-separated voxelization yields $V_{k+1}\in\R^{2\times B\times H\times W}$ with $B$ temporal bins and an $H\times W$ spatial grid. A shared encoder produces $f_E\in\R^{512}$, from which two heads predict rotational and image-plane candidates $\hat{\delta\phi}\in\R^3$ and $\hat{\delta q}\in\R^2$.

For positive camera-frame depth, define
\begin{equation}
q=\begin{bmatrix}x/z\\y/z\end{bmatrix},\quad
q_p=\begin{bmatrix}x_p/z_p\\y_p/z_p\end{bmatrix},\quad
\mathcal T(q,z)=z\begin{bmatrix}q_x\\q_y\\1\end{bmatrix}.
\label{eq:bearing}
\end{equation}
The reference pose $(R^*,t^*)$ supplies training residuals
\begin{equation}
\delta\phi^*=\Log(R_p^\top R^*)^\vee,\qquad
\delta q^*=q^*-q_p .
\label{eq:targets}
\end{equation}
Both candidates and the Scale Gate use the event-only encoder. The prior defines the residual reference and supplies the gate's motion scalars.

The projection $\pi(t)=[x/z,y/z]^\top$ relates translation to image-plane displacement through
\begin{equation}
J_\pi(t_p)=\frac{1}{z_p}
 \begin{bmatrix}1&0&-q_{x,p}\\0&1&-q_{y,p}\end{bmatrix},
\quad
\delta q\simeq\frac{\delta t_{xy}-q_p\delta z}{z_p}.
\end{equation}
Scaling $t_p$ uniformly leaves its point bearing unchanged. Transverse displacement changes that bearing, which is the translation component corrected by OA-EI; $z_p$ is retained. The chosen subspace does not exhaust depth information carried by extended-object scale and temporal geometry.

With injection weights $\alpha_R,\alpha_q\in[0,1]$, the selective update is
\begin{subequations}
\label{eq:update}
\begin{align}
R_f&=R_p\Exp([\alpha_R\hat{\delta\phi}]_\times),\label{eq:updateR}\\
q_f&=q_p+\alpha_q\hat{\delta q},\label{eq:updateq}\\
z_f&=z_p,\label{eq:updatez}\\
t_f&=\mathcal T(q_f,z_f)=z_p[q_{x,f},q_{y,f},1]^\top .\label{eq:updatet}
\end{align}
\end{subequations}
The translation change is $t_f-t_p=z_p\alpha_q[\hat{\delta q}^\top,0]^\top$, on the prior's axial-depth plane (Fig.~\ref{fig:geometry}). Injection weights determine how much of each candidate is applied.

\begin{figure*}[!t]
\centering
\includegraphics[width=\textwidth]{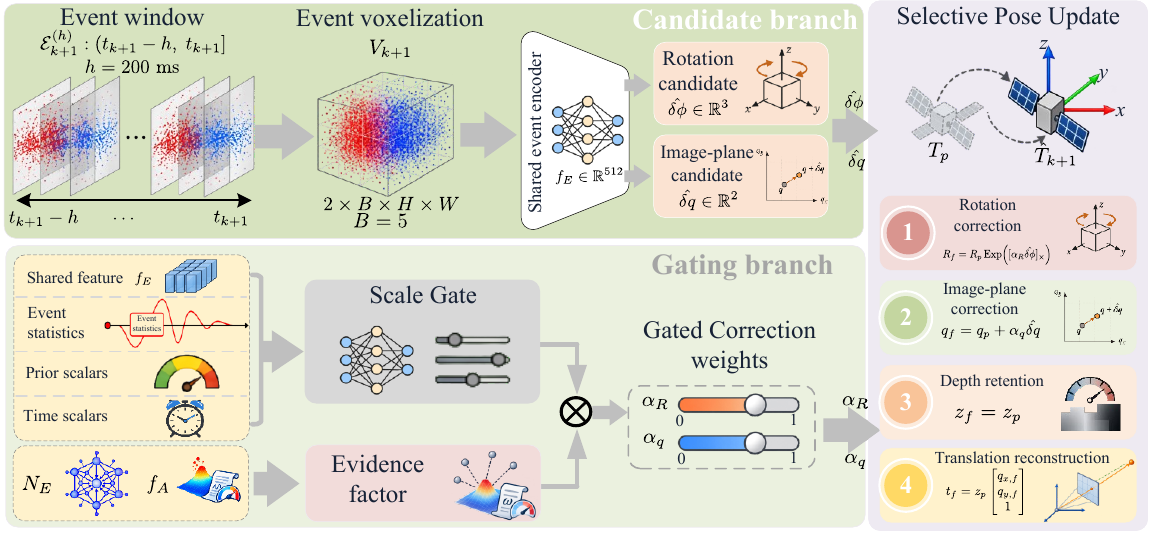}
\caption{Event candidates, gated weights, and selective pose composition. A causal $h=200$-ms window is voxelized into two polarities and $B=5$ temporal bins. The shared feature $f_E$ supplies rotation and image-plane candidates and the Scale Gate. Predicted scale coefficients are multiplied by the deterministic Evidence factor derived from $N_E$ and $f_A$ to obtain $\alpha_R,\alpha_q\in[0,1]$. The right panel applies the weighted corrections, retains $z_p$, and reconstructs $t_f$. Prior-motion and timing scalars condition the gate.}
\label{fig:eventupdate}
\end{figure*}

\subsection{Learned and Activity-Supported Injection}
Predicted candidate scales and deterministic activity support enter the update multiplicatively (Fig.~\ref{fig:eventupdate}):
\begin{equation}
\alpha_b=\alpha_{\mathrm{scale}}^b r_{\mathrm{evidence}},
\qquad b\in\{R,q\}.
\label{eq:injection}
\end{equation}
Both factors lie in $[0,1]$. The Scale Gate predicts $\alpha_{\mathrm{scale}}^R,\alpha_{\mathrm{scale}}^q$ from $f_E$ and nine causal scalars: log event count, log event rate, active-pixel fraction, spatial event entropy, polarity balance, log prior angular and translation speeds, $h$, and $\Delta t$. Training-set statistics normalize these scalars.

The Scale Gate is trained after freezing the candidate network. Targets minimize rotation geodesic error $e_R$ and translation error $e_T(t,t^*)=\|t-t^*\|_2$ along the proposed directions:
\begin{align}
\alpha_R^*&=\underset{\alpha\in[0,1]}{\arg\min}\;
 e_R(R_p\Exp([\alpha\hat{\delta\phi}]_\times),R^*),\\
\alpha_q^*&=\underset{\alpha\in[0,1]}{\arg\min}\;
 e_T(\mathcal T(q_p+\alpha\hat{\delta q},z_p),t^*).
\end{align}
Rotation uses grid search with local refinement. For $v_q=z_p\hat{\delta q}$ and $r_q=[x^*,y^*]^\top-z_pq_p$, the translation target has the analytic form
\begin{equation}
\alpha_q^*=\operatorname{clip}
 \left(\frac{v_q^\top r_q}{v_q^\top v_q},0,1\right).
\end{equation}
A masked Smooth-L1 loss fits the predicted scales, excluding near-zero candidate directions. Nonnegative coefficients preserve each candidate's direction.

For the Evidence factor, event count $N_E$ and active-pixel fraction $f_A$ are computed by summing the voxel over time and polarity. Their fifth percentiles on normal training windows set $\theta_N,\theta_A$. With $S(r)=3u^2-2u^3$ and $u=\operatorname{clip}(r,0,1)$, this count-and-coverage heuristic takes the smaller activity term:
\begin{equation}
r_{\mathrm{evidence}}=
 \min\{S(N_E/\theta_N),S(f_A/\theta_A)\}.
\label{eq:support}
\end{equation}
An empty window gives
\begin{equation}
N_E=0\ \Rightarrow\ \alpha_R=\alpha_q=0
 \ \Rightarrow\ T_f=T_p .
\label{eq:fallback}
\end{equation}
The next prediction uses the pair formed by $\hat T_k$ and the corrected pose $\hat T_{k+1}=T_f$.

\subsection{Implementation and Training}
The event representation uses $h=200$~ms, $\Delta t=100$~ms, and five temporal bins per polarity at $45\times80$ resolution. Windows end at their target timestamps, overlapping by 100~ms. The ResNet-18-style encoder outputs 512 features. Each candidate head has a $512\!\to\!256$ linear layer followed by LayerNorm, GELU, dropout of 0.1, and a three- or two-dimensional output.

Componentwise training standard deviations $s_R,s_q$ normalize the rotation and plane residuals, with floors of $10^{-4}$~rad and $10^{-5}$. The candidate loss is
\begin{equation}
\mathcal L_{\mathrm{cand}}=
 \rho((\hat{\delta\phi}-\delta\phi^*)\oslash s_R)+
 \rho((\hat{\delta q}-\delta q^*)\oslash s_q),
\end{equation}
where $\rho$ denotes Smooth-L1 loss and $\oslash$ is elementwise division. The gate combines a $512\!\to\!128$ event-feature branch and a $9\!\to\!32$ scalar branch in a $160\!\to\!128\!\to\!2$ sigmoid head. It has 87,892 parameters.

AdamW trains the candidate and gate for up to 30 epochs, at respective learning rates $3\times10^{-4}$/$10^{-3}$ and batch sizes 128/512. Both use weight decay $10^{-4}$, cosine decay, and gradient clipping at 5; seeds are 3407, 3408, and 3409. On the development partition, candidates are ranked by worst rotation/translation error ratio and then geometric mean, including an epoch-0 zero-correction candidate.

The selected candidate stays frozen during gate training. Clean windows are supplemented with voxel-level count thinning, temporal-bin permutation, and reversal; gate checkpoints minimize masked scale-prediction loss. Rotation targets use a 101-point grid and a subsequent 41-point refinement within radius 0.01. Masks exclude rotation norms $\le10^{-8}$ and squared $v_q$ norms $\le10^{-10}$. Reference pose histories are used in both training stages, with self-fed histories reserved for evaluation.

\section{Experiments}
The main comparison follows self-fed tracking over the full rollout. Update and injection ablations examine the two design choices within that protocol; contextual comparisons and embedded measurements use the resulting tracker.

\begin{figure*}[!t]
\centering
\includegraphics[width=\textwidth]{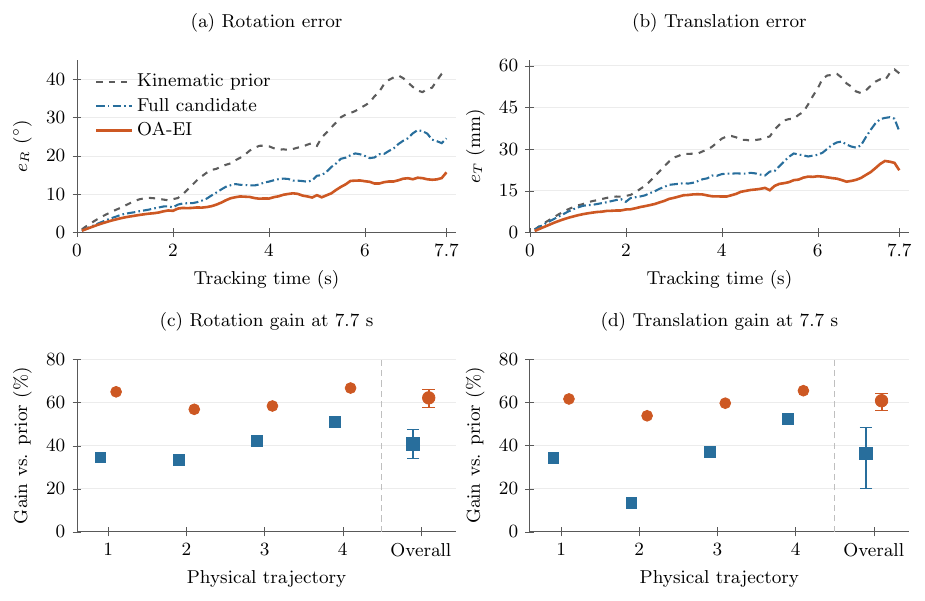}
\caption{Recursive tracking behavior. Mean rotation (a) and translation (b) errors over 0.1--7.7 s are paired with endpoint gains relative to the kinematic prior for all four physical trajectories (c,d). Blue squares denote full candidate injection and orange circles denote OA-EI. Separate overall markers show the cohort-level gain with descriptive 95\% trajectory-cluster bootstrap intervals.}
\label{fig:tracking}
\end{figure*}

\begin{table*}[!t]
\caption{Recursive Tracking Performance}
\label{tab:tracking}
\centering\footnotesize
\renewcommand{\arraystretch}{1.15}\setlength{\tabcolsep}{4pt}
\begin{tabularx}{\textwidth}{Yrrrrrrrrr}
\toprule
Tracker & \multicolumn{2}{c}{2.0 s} & \multicolumn{2}{c}{5.0 s} & \multicolumn{2}{c}{7.7 s} & \multicolumn{2}{c}{Error area} & Failure \\
\cmidrule(lr){2-3}\cmidrule(lr){4-5}\cmidrule(lr){6-7}\cmidrule(lr){8-9}
& $e_R\downarrow$ ($^\circ$) & $e_T\downarrow$ (mm) & $e_R\downarrow$ ($^\circ$) & $e_T\downarrow$ (mm) & $e_R\downarrow$ ($^\circ$) & $e_T\downarrow$ (mm) & $A_R\downarrow$ ($^\circ$) & $A_T\downarrow$ (mm) & (\%)$\downarrow$ \\
\midrule
Kinematic prior & 8.73 & 13.18 & 22.64 & 34.57 & 41.63 & 57.28 & 21.37 & 30.84 & 20.00 \\
Full candidate & 6.57 & 11.06 & 14.83 & 21.94 & 24.61 & 36.42 & 13.28 & 20.37 & 8.33 \\
\textbf{OA-EI (ours)} & \textbf{5.68} & \textbf{8.31} & \textbf{9.76} & \textbf{15.24} & \textbf{15.73} & \textbf{22.41} & \textbf{8.87} & \textbf{13.62} & \textbf{1.67} \\
\bottomrule
\end{tabularx}
\end{table*}

\subsection{Experimental Setup and Evaluation Protocol}
The FRESH \cite{jawaid2026fresh} fixed-camera Prophesee EVK4 recordings provide the reference poses and event data. The split comprises 16 training, four development, and four held-out trajectories, using an aligned $800\times720$ central crop. Method design and all parameter/checkpoint selection are confined to the training and development partitions.

Five starts are taken from each held-out trajectory. Every rollout receives two reference poses 100~ms apart, followed by 77 self-fed updates over 7.7~s; starts and target timestamps are shared across methods. The learned methods' three training seeds (3407, 3408, 3409) give 60 runs, while evaluating the deterministic prior once per start gives 20. Seed errors are averaged within each start, starts within each trajectory, and trajectories with equal weight. There is no seed averaging for the prior. The paired trajectory-cluster bootstrap intervals in Fig.~\ref{fig:tracking} use 2,000 resamples, 95\% percentile bounds, and RNG seed 3407. They describe the four physical trajectories.

We report rotation geodesic error $e_R$ ($^\circ$), translation error $e_T=\|t-t^*\|_2$ (mm), and time-normalized error area:
\begin{equation}
A_b=\frac{1}{t_N-t_1}\sum_{k=1}^{N-1}
 \frac{e_{b,k}+e_{b,k+1}}{2}\,\Delta t_k .
\label{eq:errorarea}
\end{equation}
Here $e_{b,k}=e_b(t_k)$, $b\in\{R,T\}$, $\Delta t_k=t_{k+1}-t_k$, $N=77$, and $[t_1,t_N]=[0.1,7.7]$~s. Relative gain is
\begin{equation}
G_b=100\,\frac{\bar e_{b,\mathrm{prior}}-\bar e_{b,\mathrm{method}}}
 {\bar e_{b,\mathrm{prior}}}.
\label{eq:gain}
\end{equation}
The first missing, non-finite, or invalid pose output marks a rollout as failed for its remaining duration. The last valid pose is then held, and its errors against subsequent target references remain in the endpoint, area, and curve statistics. Failure rates use 20 prior rollouts or 60 learned-method runs. Bold entries denote the lowest table errors or failure rates.

\subsection{Recursive Tracking Performance}
\label{sec:tracking}
Constant-twist prediction is evaluated alongside full candidate injection with unit weights and OA-EI's learned scaling and activity support. Full injection and OA-EI use the same rotational and image-plane candidates. After the common initialization, each tracker recursively feeds back its own pose estimates.

Table~\ref{tab:tracking} shows the lowest errors for OA-EI at all reported horizons. Its 7.7~s errors are $15.73^\circ$/22.41~mm, reductions of 62.2\%/60.9\% from the prior. The interval-wide error areas also decrease, by 58.5\%/55.8\%.

There are 4/20 failed prior rollouts and 5/60 failed full-injection runs, compared with 1/60 for OA-EI. The temporal curves in Fig.~\ref{fig:tracking} accompany positive OA-EI endpoint gains on all four trajectories.

\subsection{Matched State-Update Ablation}
\label{sec:matched}
The state-update alternatives are $R+\mathrm{XYZ}$ Cartesian residuals, $R+q+z$ with an additional depth residual, and unified $\SE(3)$ updates, evaluated against OA-EI's $R+q$ interface. They share the encoder, data, training budget, seeds, event windows, initialization, development selection, and scaling/activity protocol. Each has a separate scale fit under the same geometric supervision. Head sizes depend on parameterization and are reported to the nearest thousand in Table~\ref{tab:matched}, excluding the shared encoder and Scale Gate. The 7.7~s endpoint metrics include lateral error $e_{xy}=\|[x-x^*,y-y^*]^\top\|_2$ and axial error $e_z=|z-z^*|$.

The lowest $e_R$, $e_T$, $e_{xy}$, and failure rate belong to OA-EI, but the axial-error ranking differs. Adding depth correction in $R+q+z$ lowers axial error to 6.71~mm from OA-EI's 7.26~mm; total translation error rises to 30.42~mm from 22.41~mm. The $R+q$ interface therefore favors overall pose accuracy. Retaining $z_p$ in an update still allows its future propagation to change, since corrected poses enter the subsequent prediction history.

\begin{table}[!h]
\caption{Matched State-Update Ablation}
\label{tab:matched}
\centering\footnotesize
\renewcommand{\arraystretch}{1.15}\setlength{\tabcolsep}{2pt}
\begin{tabularx}{\columnwidth}{Yrrrrrr}
\toprule
State update & $e_R$ & $e_T$ & $e_{xy}$ & $e_z$ & Head & Fail.\\
& ($^\circ$) & (mm) & (mm) & (mm) & (k) & (\%)\\
\midrule
Kinematic prior & 41.63 & 57.28 & 55.21 & 14.23 & --- & 20.00\\
$R+\mathrm{XYZ}$ & 24.38 & 33.71 & 32.68 & 7.84 & 265 & 6.67\\
$R+q+z$ & 20.76 & 30.42 & 29.54 & \textbf{6.71} & 265 & 5.00\\
Unified $\SE(3)$ & 27.34 & 37.65 & 36.31 & 8.83 & 133 & 10.00\\
\textbf{OA-EI:} $R+q$ & \textbf{15.73} & \textbf{22.41} & \textbf{21.08} & 7.26 & 265 & \textbf{1.67}\\
\bottomrule
\end{tabularx}
\end{table}

\subsection{Injection Scaling and Sparse-Event Robustness}
\label{sec:weightanalysis}
Learned scaling and activity support are independently enabled in a $2\times2$ ablation; a disabled factor is set to one. The candidate network/checkpoint is shared and frozen, as are gate weights, normalizers, and support thresholds. Tests retain 100\%, 10\%, or 1\% of raw events. Within each thinning replicate, a global event-ID mask is reused across variants and overlapping windows, and voxels and window statistics are recomputed. Pairing uses common starts and thinning seeds.

Either factor alone reduces errors in Table~\ref{tab:injection}; using both gives the lowest errors at every retention level. The 1\% case separates full injection from activity-supported scaling: full injection exceeds the prior's errors, while OA-EI reaches $27.28^\circ$/38.64~mm. Full injection, scales only, activity support only, and OA-EI have 17/60, 11/60, 8/60, and 3/60 failed runs, respectively.

\begin{table}[!h]
\caption{Scaling and Activity Support Under Event Sparsification: $e_R/e_T$ at 7.7 s ($^\circ$/mm)}
\label{tab:injection}
\centering\footnotesize
\renewcommand{\arraystretch}{1.15}\setlength{\tabcolsep}{3pt}
\begin{tabularx}{\columnwidth}{YYrrr}
\toprule
Scale Gate & Activity factor & \shortstack{100\%\\events} & \shortstack{10\%\\events} & \shortstack{1\%\\events} \\
\midrule
$\times$ & $\times$ & 24.61 / 36.42 & 32.17 / 44.08 & 45.74 / 62.31 \\
$\checkmark$ & $\times$ & 20.93 / 30.58 & 27.64 / 38.21 & 39.46 / 53.77 \\
$\times$ & $\checkmark$ & 18.84 / 27.16 & 23.38 / 32.74 & 32.71 / 47.35 \\
$\checkmark$ & $\checkmark$ & \textbf{15.73 / 22.41} & \textbf{19.11 / 28.06} & \textbf{27.28 / 38.64} \\
\bottomrule
\end{tabularx}
\end{table}

\subsection{Contextual Comparison with Existing Pipelines}
Table~\ref{tab:published} contains our reproduced results for Keypoint--PnP \cite{jawaid2026fresh}, PASAN+IRIS \cite{zeng2026smoothing}, and cascaded pose \cite{zhang2026cascaded}. Keypoint geometry, temporal smoothing, and pose optimization distinguish these pipelines from OA-EI's recursive correction. The comparison is contextual and system-level, with different input and temporal mechanisms.

Each reproduced baseline follows the three-seed, 60-run protocol in Section~IV-A. Best checkpoints are selected on the development partition; the held-out cohort, starts, timestamps, error definitions, and failure handling are shared across pipelines.

The best-performing baseline, cascaded pose, reaches $22.68^\circ$/34.37~mm at 7.7~s and fails in 4/60 runs. OA-EI reaches $15.73^\circ$/22.41~mm with 1/60 failures; it also has the lowest errors at the earlier horizon and the lowest overall failure rate.

\begin{table}[!h]
\caption{Contextual Pipeline Comparison }
\label{tab:published}
\centering\footnotesize
\renewcommand{\arraystretch}{1.15}\setlength{\tabcolsep}{2pt}
\begin{tabularx}{\columnwidth}{Yrrrrr}
\toprule
Pipeline & \multicolumn{2}{c}{2.0 s} & \multicolumn{2}{c}{7.7 s} & Fail.\\
\cmidrule(lr){2-3}\cmidrule(lr){4-5}
& $e_R$ ($^\circ$) & $e_T$ (mm) & $e_R$ ($^\circ$) & $e_T$ (mm) & (\%)\\
\midrule
Keypoint--PnP \cite{jawaid2026fresh} & 9.24 & 13.67 & 37.61 & 53.42 & 18.33\\
PASAN+IRIS \cite{zeng2026smoothing} & 7.31 & 12.28 & 30.47 & 42.66 & 11.67\\
Cascaded pose \cite{zhang2026cascaded} & 6.72 & 10.26 & 22.68 & 34.37 & 6.67\\
\textbf{OA-EI (ours)} & \textbf{5.68} & \textbf{8.31} & \textbf{15.73} & \textbf{22.41} & \textbf{1.67}\\
\bottomrule
\end{tabularx}
\end{table}

\subsection{Embedded Processing Cost}
Complexity, memory, and median/P95 latency on RTX 4090 and Jetson Orin Nano are listed in Table~\ref{tab:compute}. The timed in-memory update covers voxelization, propagation, candidate/scale inference, and pose composition. Sensor acquisition and online queue/slicing are outside this timing boundary. RTX profiling uses 32 development windows, each measured five times.
On Orin, the in-memory median is 44.56~ms and P95 is 58.77~ms, both within the 100-ms budget for the measured processing segment.

\begin{table}[!h]
\caption{Embedded Processing Cost}
\label{tab:compute}
\centering\footnotesize
\renewcommand{\arraystretch}{1.1}\setlength{\tabcolsep}{4pt}
\begin{tabularx}{\columnwidth}{Ycc}
\toprule
Metric & RTX 4090 & Orin Nano\\
\midrule
Parameters (M) & 11.55 & 11.55\\
MACs/update (M) & 177.84 & 177.84\\
Network latency (ms) & 3.19/3.22 & 13.37/17.84\\
In-memory update (ms) & 16.61/19.68 & 44.56/58.77\\
Peak memory (MB) & 85.2 & 540\\
\bottomrule
\end{tabularx}
\end{table}

\section{Conclusion}
OA-EI uses event observations to refine rotation and normalized image-plane translation within an initialized spacecraft tracker, while depth follows the kinematic prediction. Learned scaling and event-activity support control the corrections that enter the next pose history. With no events, the prior is preserved exactly.

On the evaluated FRESH cohort, 7.7-s rotation and translation errors are reduced by 62.2\% and 60.9\% relative to kinematic prediction, and failure rates decrease from 20.0\% to 1.67\%. The state-update comparison favors rotation--bearing correction for overall pose accuracy, even though adding a depth residual gives slightly lower axial error at the cost of larger total translation error. The combined Scale Gate and activity support outperform full injection at full, 10\%, and 1\% event retention.

These results concern fixed-camera laboratory sequences initialized with two reference poses, followed by recursive estimates. Estimator-derived initialization and independently captured sequences remain the next evaluation settings. Deployment evaluation will also extend the measured in-memory latency to complete sensor-to-output processing.

\bibliographystyle{IEEEtran}
\bibliography{references}

\begin{IEEEbiography}
[{\includegraphics[
    width=1in,
    height=1.25in,
    keepaspectratio
]{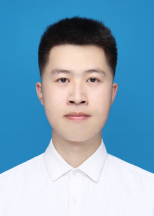}}]
{Yonglong Zhang}

received the B.S. degree in mechatronic engineering from Harbin Institute of Technology, Harbin, China, in 2024. He is currently pursuing the master's degree with the School of Mechatronics Engineering, Harbin Institute of Technology, Harbin, China. His research interests include robot pose estimation and robotic manipulation.

\end{IEEEbiography}

\begin{IEEEbiography}
[{\includegraphics[
    width=1in,
    height=1.25in,
    keepaspectratio
]{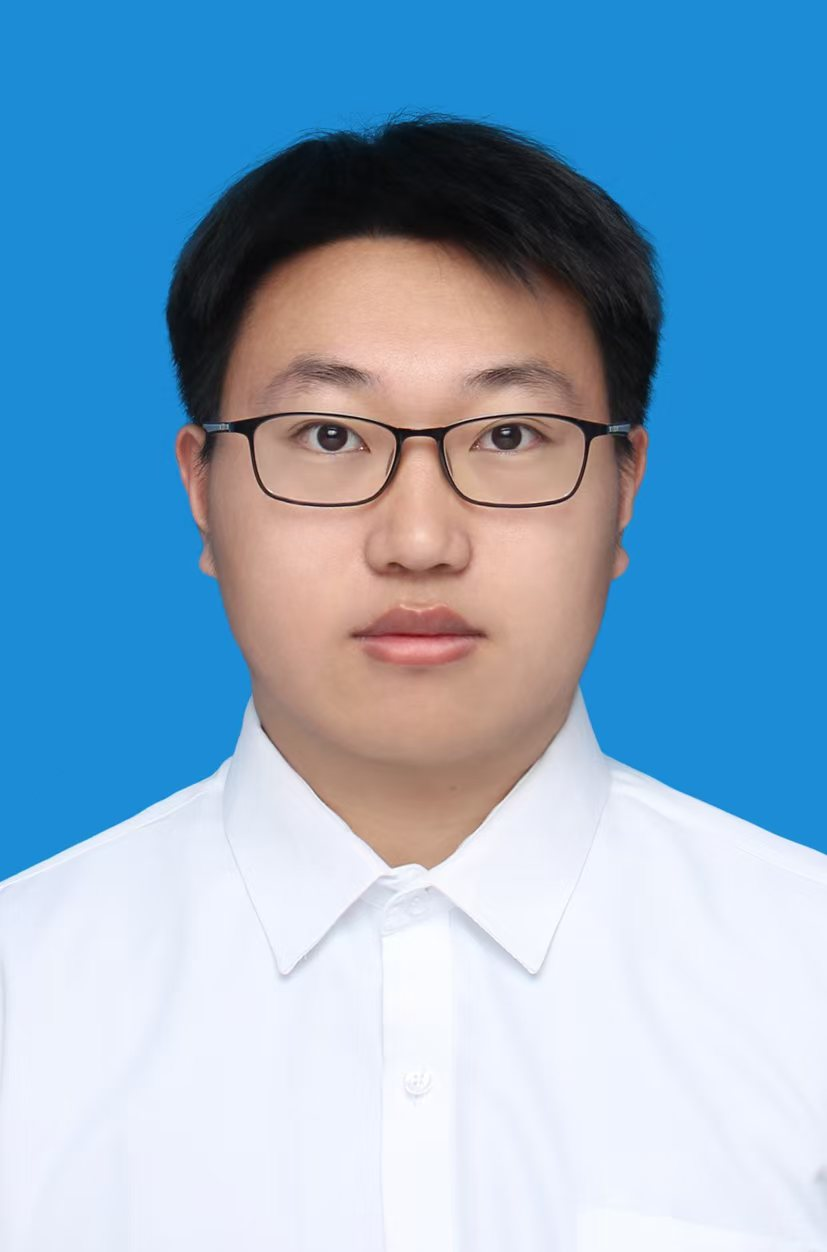}}]
{Guanning Yao}

received the B.S. degree in robotics engineering from Harbin Institute of Technology, Harbin, China, in 2025. He is currently pursuing the master's degree with the School of Mechatronics Engineering, Harbin Institute of Technology, Harbin, China. His research interests include low-disturbance control of space robots and space robotic manipulation.

\end{IEEEbiography}

\begin{IEEEbiography}
[{\includegraphics[
    width=1in,
    height=1.25in,
    keepaspectratio
]{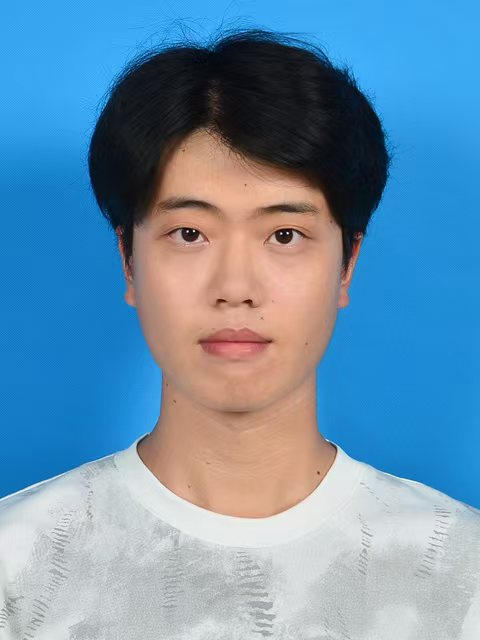}}]
{Shuo Zhang}

received the B.S. degree in robotics engineering from Harbin Institute of Technology, Harbin, China, in 2025. He is currently pursuing the master's degree with the School of Mechatronics Engineering, Harbin Institute of Technology, Harbin, China. His research interests include robot reinforcement learning, robotic manipulation, and robot pose estimation.

\end{IEEEbiography}

\begin{IEEEbiography}
[{\includegraphics[
    width=1in,
    height=1.25in,
    keepaspectratio
]{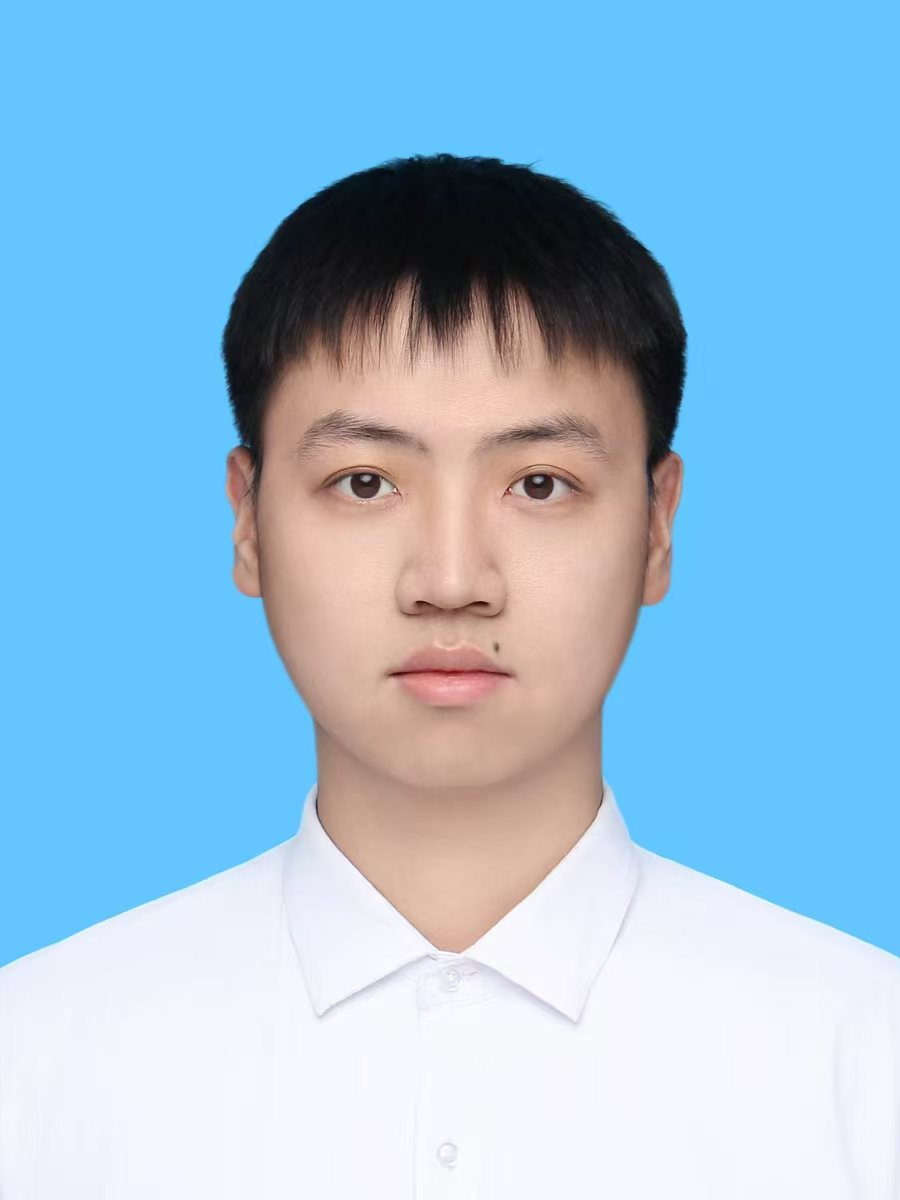}}]
{Yifan Yang}

received the B.S. degree in mechanical design, manufacturing, and automation from Harbin Institute of Technology, Harbin, China, in 2025. He is currently pursuing the master's degree with the School of Mechatronics Engineering, Harbin Institute of Technology, Harbin, China. His research interests include robot pose estimation and robotic manipulation.

\end{IEEEbiography}

\begin{IEEEbiography}
[{\includegraphics[
    width=1in,
    height=1.25in,
    keepaspectratio
]{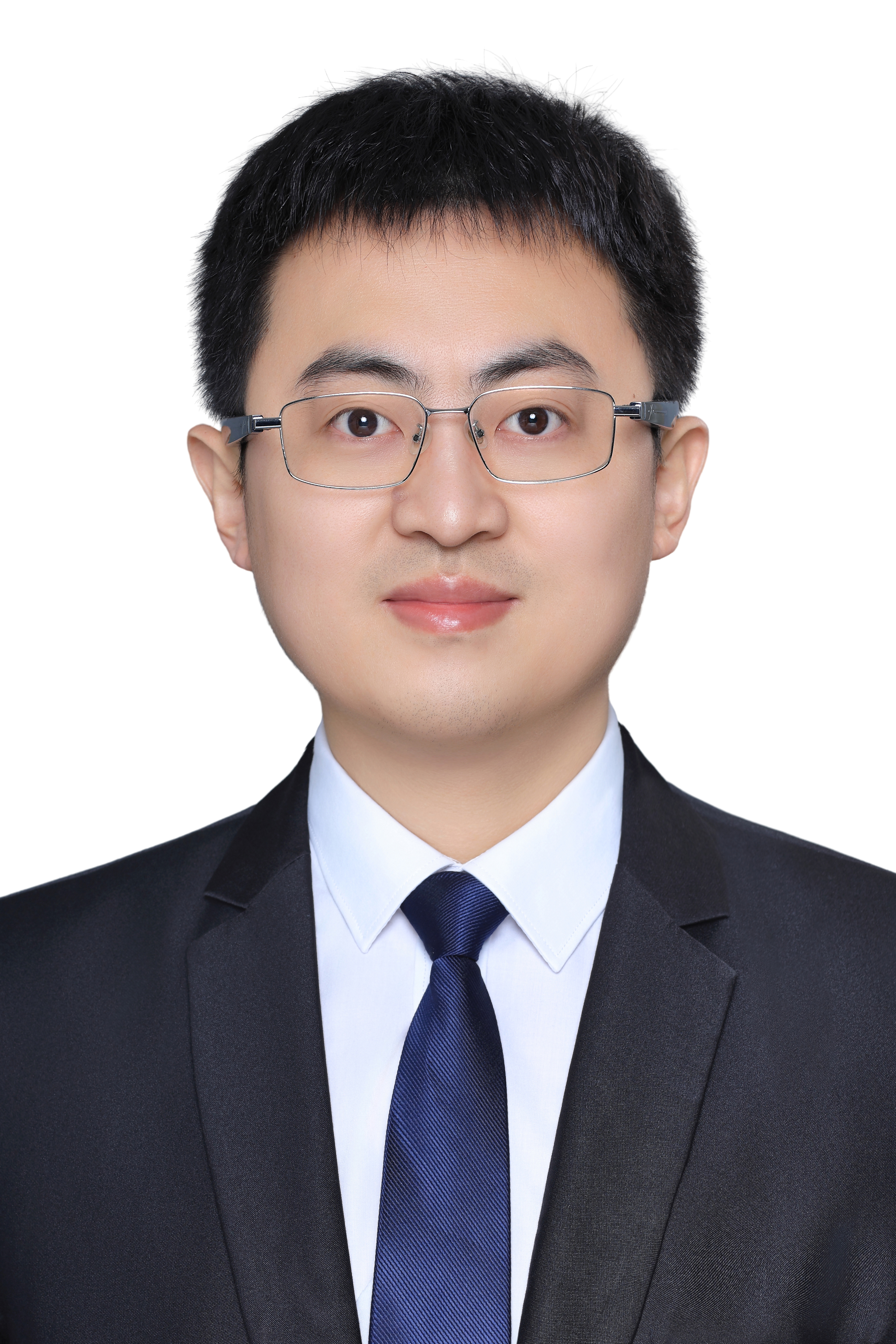}}]
{Yang Liu}

received the B.S., M.S., and Ph.D. degrees from Harbin Institute of Technology, Harbin, China, in 2013, 2015, and 2020, respectively. He is currently an Associate Professor with Harbin Institute of Technology, Harbin, China, and also serves as the Deputy Secretary of the Youth League Committee. His research interests include space robotics, humanoid robots, and computer vision. He has authored or coauthored 35 scientific papers, holds 16 invention patents, and has contributed to two textbooks. He has been selected for the Youth Talent Support Program and recognized as an Outstanding Youth of Heilongjiang Province.

\end{IEEEbiography}

\begin{IEEEbiography}
[{\includegraphics[
    width=1in,
    height=1.25in,
    keepaspectratio
]{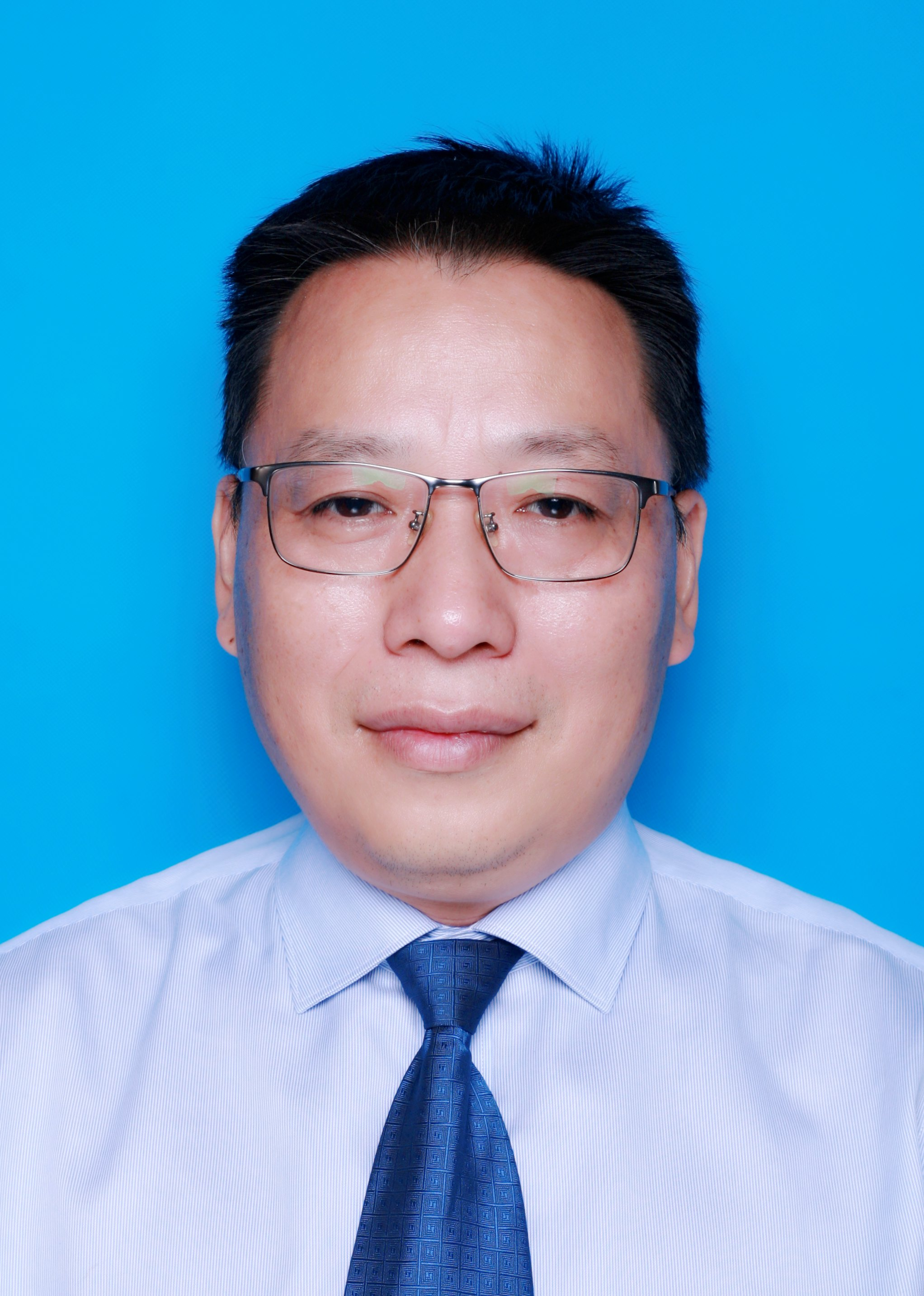}}]
{Zongwu Xie}

received the B.S. degree in electrical engineering and automation from Harbin University of Science and Technology, Harbin, China, in 1996, and the M.S. and Ph.D. degrees in mechanical engineering from Harbin Institute of Technology, Harbin, China, in 2000 and 2003, respectively.

\end{IEEEbiography}

\end{document}